\documentclass{article}

\usepackage{graphicx} 
\usepackage{subfigure} 

\usepackage{natbib}

\usepackage{algorithm}
\usepackage{algorithmic}
\usepackage{amsmath,amssymb}
\newtheorem{thm}{Theorem}[section]
\newtheorem{lem}[thm]{Lemma} 

\usepackage{hyperref}

\usepackage[accepted]{icml2012}

\icmltitlerunning{Unsupervised pre-training helps to conserve views from input distribution}

\begin{document} 

\twocolumn[
\icmltitle{Unsupervised pre-training helps to conserve views from input distribution}

\icmlauthor{Nicolas Pinchaud}{nicolas.pinchaud@gmail.com}
\icmladdress{}
\icmlauthor{}{}
\icmladdress{}

\icmlkeywords{machine learning, ICML}

\vskip 0.12in
]

\begin{abstract}

We investigate the effects of the unsupervised pre-training method under the perspective of information theory. If the input distribution displays multiple views of the supervision, then unsupervised pre-training allows to learn hierarchical representation which communicates these views across layers, while disentangling the supervision. Disentanglement of supervision leads learned features to be independent conditionally to the label. In case of binary features, we show that conditional independence allows to extract label's information with a linear model and therefore helps to solve under-fitting. We suppose that representations displaying multiple views help to solve over-fitting because each view provides information that helps to reduce model's variance. We propose a practical method to measure both disentanglement of supervision and quantity of views within a binary representation. We show that unsupervised pre-training helps to conserve views from input distribution, whereas representations learned using supervised models disregard most of them.
\end{abstract} 

\section{Introduction}
\label{introduction}

In the context of classification problem, we want to learn a conditional distribution $P(Y|X)$ of two random variables : $X$ the input vector, and $Y$ the label. To solve this, a data set $\mathcal{D}$ of samples from the joint distribution $P(X,Y)$ is provided. The goal is to learn a model $P(Y|X,\theta)$ of the true distribution, where $\theta$ represents its parameters. The deep learning approach proposes to learn an intermediate representation $H$ of $X$ using a hierarchy $H^{(1)},...,H^{(L-1)},H^{(L)}$ of random vectors, where $H=H^{(L)}$, such that we have the following factorization :
\begin{equation}
\label{deep}
\begin{array}{r}
P(H^{(1)},...,H^{(L-1)},H|X) =\\ P(H|H^{(L-1)})\times...\times P(H^{(1)}|X)
\end{array}
\end{equation}
This form allows to easily and efficiently sample $P(H|X)$ by successively sampling each factor. Using the intermediate representation $H$ may allow to easily learn a model of $P(Y|H)$ and then get $P(Y|X)$ by composition : $P(Y,H|X)=P(Y|H)P(H|X)$. This process may be more successful than directly learn a model of $P(Y|X)$, especially if we know how to get good representations. A popular method, introduced by \cite{DBN} and \cite{NIPS2006_739}, is to learn $P(H|X)$ by greedily stacking factors $P(H^{(l+1)}|H^{(l)})$ learned using unsupervised models like RBMs \cite{EFH} or auto-encoders \cite{Vincent08extractingand} or variations of them \cite{ICML2011Rifai_455}\cite{ranzato10}. Why does this method give rise to distribution $H$ that helps to solve the supervised problem ? The reasons remain mainly unclear. It has been shown that stacking RBMs is equivalent to learn a deep generative model of data \cite{DBN}. The intuition is that knowledge of how the data behave allows to deduce the label easily since it is just one of the aspects of this behavior.  Other works suggest that this method act as a regularization helping to find model's parameters that generalize well \cite{Erhan-aistats-2010}, or show that it helps to learn invariant features \cite{Goodfellow+etal09}. We propose an approach by an information theoretic \cite{shannon} analysis of features obtained with unsupervised pre-training. Deep learning usually deals with input distributions that are high dimensional random vectors. These distributions are likely to display information about supervision in multiple ways. For example there may exist various subsets of components from the vector that provide full information about labels. These views can be helpful for generalization as they provide information that may help to determine the parameters of the model. Experiments show that supervised approaches naively try to disentangle the supervision without concern about preservation of these views. On the contrary, unsupervised pre-training does both, disentanglement of supervision and preservation of views.
Disentanglement of supervision is related to features being independent conditionally to the supervision $Y$. If $H$ is binary, we show that a conditional independence with $Y$ enables to model $P(Y|H)$ with a linear model.

%

The paper is organized as follows. We first define our framework in section \ref{framework}. Section \ref{disentanglement} relates conditional independence with disentanglement of supervision in the case of binary features. In section \ref{multiple_views} we propose to define a measure of relevance of views contained in representations using information interaction. The section \ref{profile} defines a practical method to measure both conditional independence of components and view relevance. Finally, experiments are presented in section \ref{experiments}.

%
%
%
\section{Framework and Notations}
\label{framework}
A deep representation \eqref{deep} is trained by greedily stacking simpler learning modules. We start by learning the parameters $\theta^{(1)}$ of the first one $P(H^{(1)}|X,\theta^{(1)})$, which is then used to train the next one $P(H^{(2)}|H^{(1)},\theta^{(2)})$, and so on. We abstract the notation of a module with the parametrized distribution $P(B|V,\theta)$, where $(B,V)$ is instantiated with corresponding pair\footnote{With $H^{(0)}=X$} $(H^{(l)},H^{(l-1)})$. We suppose that layers are binary, $B \in \{0,1\}^m$. We note $B_i$, the $i^{th}$ component of $B$. We note $\Pi_n$ the set of subsets of components of $B$ of size $n$, and $\Pi^i_n$ the set of subsets of components of $B$ of size $n$ that do not contain $B_i$. We note $\mathbf{H}(.)$ and $\mathbf{I}(.,.)$ the Shannon entropy and the mutual information. We suppose that the inference of $B$ is easy by assuming that :
\begin{equation}\label{inde}
P(B|V,\theta) = \prod_i P(B_i|V,\theta)
\end{equation}
We suppose that $Y$ is a discrete random variable.
\section{Supervision disentanglement as conditional independence}
\label{disentanglement}
In this section we relate disentanglement of $Y$ with independence of components of $B$ conditionally to $Y$ by pointing out that the latter enables classes to be linearly separable. This is shown by the following theorem :
\begin{thm}\label{thm1}
Let $Y$ be a discrete random variable taking values in $\mathcal{Y}$, let $B$ be a binary random vector taking values in $\{0,1\}^m$.

If $\mathbf{H}(Y|B)=0$, and if components of $B$ are independent conditionally to $Y$, then for all classes $y_n \in \mathcal{Y}$ so that  $P(y_n)>0$, there exists a vector $\mu^{(n)} \in \mathcal{R}^{m+1}$ such that $\forall b, P(B=b)\neq 0$ we have :
\[
(b_0,...,b_{m-1},1).\mu^{(n)} > 0 \Leftrightarrow P(y_n|b)=1
\]
\end{thm}

The underlying idea of the proof is that if we have conditional independence, then for each class $y_n$ values of $B|y_n$ are restricted to a sub-hypercube of $\{0,1\}^m$ such that there is no intersection between sub-hypercubes of different classes. This allows to find hyper-planes that separate them.  

Before proving the theorem, we introduce several lemmas. We suppose that hypotheses $\mathbf{H}(Y|B)=0$ and $P(B|Y)=\prod_i P(B_i|Y)$ (conditional independence of components) are satisfied. 

Let $y_n \in \mathcal{Y}$ so that $P(Y=y_n)>0$, we note :
\small
\[\mathcal{B}_n = \{b \in \{0,1\}^m, P(b|y_n)\neq 0\}\]
\normalsize

\begin{lem}\label{0} Let $y_n \in \mathcal{Y}$ so that $P(Y=y_n)>0$, let $y_l \in \mathcal{Y}$ so that $P(Y=y_l)>0$ and $y_l\neq y_n$, then $\mathcal{B}_n \cap \mathcal{B}_l = \emptyset$ \end{lem}

\emph{Proof by contradiction :} Suppose $\exists b \in \mathcal{B}_n \cap \mathcal{B}_l$, then we have $P(b|y_n)>0$ and $P(b|y_l)>0$. By the Bayes theorem and because $P(b)>0$, $P(y_n)>0$, $P(y_l)>0$, we can write :
 \[  
  \begin{array}{l}
    
    P(b|y_n)>0 \Rightarrow P(y_n|b)>0\\
    P(b|y_l)>0 \Rightarrow P(y_l|b)>0\\
  \end{array} 
\]
Since we suppose that $\mathbf{H}(Y|B)=0$, then $\exists ! y, P(y|b)=1$, so $\forall \tilde{y}\neq y, P(\tilde{y}|b)=0$. We deduce that $P(y_n|b)=P(y_l|b)=1$, and then $y_n=y_l$ which contradicts the hypothesis $y_n \neq y_l$ $\square$

Let $\mathcal{B}=\bigcup_{y_n \in \mathcal{Y},P(y_n)>0}\limits \mathcal{B}_n$

\begin{lem}\label{0.5}
We have \[\mathcal{B} = \{b \in \{0,1\}^m, P(B=b)>0\}\]
\end{lem}
\emph{Proof :} Let $\hat{b} \in \mathcal{B}$, then $\exists y_n\in \mathcal{Y}, P(y_n)>0 , \hat{b} \in \mathcal{B}_n$. Therefore $P(\hat{b}|y_n)>0$, which implies that $p(\hat{b})>0$, then $\hat{b}\in  \{b \in \{0,1\}^m, P(B=b)>0\}$.

Let $\hat{b} \in \{b \in \{0,1\}^m, P(B=b)>0\}$, then since $\mathbf{H}(Y|B)=0$, $\exists ! y_n, P(y_n|\hat{b})=1$, and knowing that $P(\hat{b})>0$, the Bayes theorem allows us to write $P(\hat{b}|y_n)>0$, then $\hat{b} \in \mathcal{B}_n$ $\square$

Let $y_n \in \mathcal{Y}$ so that $P(Y=y_n)>0$, we consider $\varphi^{(n)} \in \{-1,0,1\}^m$ so that
\[
  \varphi^{(n)}_i = \left\{ 
  \begin{array}{l l}
    0 & \quad \text{if $P(B_i=1|y_n)\in ]0,1[$}\\
    -1 & \quad \text{if $P(B_i=1|y_n)=0$}\\
    1 & \quad \text{if $P(B_i=1|y_n)=1$}\\
  \end{array} \right.
\]

%
%
%

\begin{lem}\label{1} Let $y_n \in \mathcal{Y}$ so that $P(Y=y_n)>0$, let $y_l \in \mathcal{Y}$ so that $P(Y=y_l)>0$ and $y_l\neq y_n$, then $\exists i, \varphi^{(n)}_i=-\varphi^{(l)}_i\neq 0 $ \end{lem}

\emph{Proof by contradiction :} If the lemma is false, for all $i$, we have one of the following case :
\[
 \left\{
 \begin{array}{l}
\varphi^{(n)}_i=\varphi^{(l)}_i\neq 0 \\

\varphi^{(n)}_i=\varphi^{(l)}_i= 0 \\

\varphi^{(n)}_i\neq 0 \text{ and } \varphi^{(l)}_i= 0 \\

\varphi^{(n)}_i = 0 \text{ and } \varphi^{(l)}_i \neq 0
\end{array}
\right.
\]
Let $b\in \{0,1\}^m$ such that $\forall i :$

\[
  b_i = \left\{ 
  \begin{array}{l l}
    \frac{\varphi^{(n)}_i+1}{2} & \quad \text{if $\varphi^{(n)}_i=\varphi^{(l)}_i\neq 0$}\\
    0 & \quad \text{if $\varphi^{(n)}_i=\varphi^{(l)}_i= 0$}\\
    \frac{\varphi^{(n)}_i+1}{2} & \quad \text{if $\varphi^{(n)}_i\neq 0$ and $\varphi^{(l)}_i= 0$}\\
     \frac{\varphi^{(l)}_i+1}{2} & \quad \text{if $\varphi^{(n)}_i= 0$ and $\varphi^{(l)}_i \neq 0$}\\
  \end{array} \right.
\]

%
%
%
%
%
%

By construction we have $\forall i, P(B_i=b_i|y_n)>0$ and $P(B_i=b_i|y_l)>0$, and the conditional independence gives us $ P(B=b|y_n)>0$ and $P(B=b|y_l)>0$, therefore $b\in \mathcal{B}^{(n)}\cap\mathcal{B}^{(l)}$ which is in contradiction with the lemma \ref{0} $\square$

Let $y_n \in\mathcal{Y}, P(y_n)>0$, $ b \in \mathcal{B}_n$, we define a vector $S^{(n,b)} \in \mathcal{R}^m$ such that :

$\forall i,S^{(n,b)}_i=(2b_i-1)\varphi^{(n)}_i$

\begin{lem}\label{2} Let $y_n \in \mathcal{Y}, P(y_n)>0$, $ b \in \mathcal{B}_n$, we have $\sum_i S^{(n,b)}_i = (2b-1)^T.\varphi^{(n)}=(\varphi^{(n)})^T.\varphi^{(n)}$\end{lem}

\emph{Proof :} It suffices to see that $\varphi^{(n)}_i\neq 0 $ then $2b_i-1=\varphi^{(n)}_i$ $\square$

\begin{lem}\label{3}  Let $y_n \in\mathcal{Y}, P(y_n)>0$, $ b \in \mathcal{B}_n$, let $\tilde{b} \in \mathcal{B}$ so that $\tilde{b} \notin \mathcal{B}_n$, then $\sum_i S^{(n,\tilde{b})}_i + 2 \leq \sum_i S^{(n,b)}_i$\end{lem}

\emph{Proof :} If $b \in \mathcal{B}$ and $\tilde{b} \notin \mathcal{B}_n$, then $\exists ! y_l, P(y_l)>0, y_l\neq y_n, \tilde{b} \in \mathcal{B}_l$.

Let $0\leq i <m$, necessarily $S^{(n,\tilde{b})}_i\leq 1$, we distinguish three possible cases depending on $\varphi^{(n)}_i$ :

if $\varphi^{(n)}_i=0$ then $S^{(n,\tilde{b})}_i=S^{(n,b)}_i=0$

if $\varphi^{(n)}_i=1$ then $b_i=1$ and $S^{(n,b)}_i=1$ so $S^{(n,\tilde{b})}_i \leq S^{(n,b)}_i$

if $\varphi^{(n)}_i=-1$ then $b_i=0$ and $S^{(n,b)}_i=1$ so $S^{(n,\tilde{b})}_i \leq S^{(n,b)}_i$

we deduce that $S^{(n,\tilde{b})}_i \leq S^{(n,b)}_i$ and that $S^{(n,b)}_i=1 \Longleftrightarrow \varphi^{(n)}_i\neq 0$

By the lemma \ref{1}, $\exists j$ so that $ \tilde{b_j} \neq  b_j$ and $\varphi^{(n)}_j\neq 0$, we have then $S^{(n,b)}_j=1$ and $S^{(n,\tilde{b})}_j=-1$ and then $S^{(n,\tilde{b})}_j+2\leq S^{(n,b)}_j$.

We conclude that $\sum_i S^{(n,\tilde{b})}_i + 2 \leq \sum_i S^{(n,b)}_i$ $\square$

Let $y_n \in \mathcal{Y}, P(y_n)>0$, we define the vector $\mu^{(n)} \in \mathcal{R}^{m+1}$ so that
\[
  \mu^{(n)}_i = \left\{ 
  \begin{array}{l l}
    2\varphi^{(n)}_i & \quad \text{if $0\leq i<m $}\\
    1-\sum_i [(\varphi^{(n)}_i)^2+\varphi^{(n)}_i] & \quad \text{if $i=m$}\\
  \end{array} \right.
\]

Note that for every $b\in \{0,1\}^m$, $$(b_0,...,b_{m-1},1).\mu^{(n)} = \sum_i S^{(n,b)}_i - (\varphi^{(n)})^T.\varphi^{(n)} + 1$$

\begin{lem}\label{4} Let $y_n \in \mathcal{Y}, P(y_n)>0$, let $b\in \mathcal{B}$,

If $b \in \mathcal{B}_n$, then $(b_0,...,b_{m-1},1).\mu^{(n)} > 0$

If $b \notin \mathcal{B}_n$, then $(b_0,...,b_{m-1},1).\mu^{(n)} < 0$ \end{lem}

\emph{Proof :} If $b \in \mathcal{B}_n$, then by lemma \ref{2}, we have $(b_0,...,b_{m-1},1).\mu^{(n)} = 1$

If  $b \notin \mathcal{B}_n$, then by lemma \ref{2} and lemma \ref{3}, we have
$(b_0,...,b_{m-1},1).\mu^{(n)} \leq -1$ $\square$

We can now prove the theorem \ref{thm1} :

Let $y_n \in \mathcal{Y}$ so that $P(y_n)>0$, let $b$ so that $P(B=b)>0$.

\emph{Proof of necessity :} We suppose that $(b_0,...,b_{m-1},1).\mu^{(n)} > 0$.

Since $P(b)>0$, then $b\in \mathcal{B}$, therefore, by lemma \ref{0} and \ref{0.5}, $\exists ! y_l, P(y_l)>0, b\in \mathcal{B}_l$.

If $y_l\neq y_n$, then $b\notin \mathcal{B}_n$ and the lemma \ref{4} tells us that $(b_0,...,b_{m-1},1).\mu^{(n)} < 0$ which contradicts the hypothesis made at the beginning of the proof. We have then $y_l=y_n$. Since $y_l=y_n$, then $b \in \mathcal{B}_n$, so $P(b|y_n)>0$. Since $\mathbf{H}(Y|B)=0$, then necessarily $P(y_n|b)=1$. $\square$

\emph{Proof of sufficiency :} If $P(y_n|b)=1$, then since $P(b)>0$, $p(y_n)>0$ and with the use of Bayes theorem, we can say that $P(b|y_n)>0$, then $b \in \mathcal{B}_n$. If $b \in \mathcal{B}_n$, then the lemma \ref{4} ensures that $(b_0,...,b_{m-1},1).\mu^{(n)} > 0$ $\square$
\subsection{A measure of conditional dependency}

We propose to measure the conditional dependency between a subset $\mathcal{I}$ of components of $B$ with the following function :
$$d_B(\mathcal{I}) := \frac{1}{|\mathcal{I}|}\left(\sum_{B_i \in \mathcal{I}}H(B_i|Y) - H(\mathcal{I}|Y) \right)$$
$d_B(\mathcal{I})$ is a normalized version of conditional total correlation between variables in $\mathcal{I}$, this normalization allows to compare the measurements between subsets of different sizes. $d_B(\mathcal{I})=0$ if and only if components in $\mathcal{I}$ are independent conditionally to $Y$. We note $D_B(n)$ the average value of $d_B$ over subsets of size $n$ :
$$
D_B(n) := \frac{1}{|\Pi_n|}\sum_{\mathcal{I} \in \Pi_n}d_B(\mathcal{I})
$$

\section{Representations displaying multiple views}
\label{multiple_views}

If the input distribution $X$ has high dimensionality, it is likely that it expresses the information about $Y$ in multiple ways. For example, in a case of face recognition task, it is likely that observation of only halves of faces should be enough to deduce their label. This distribution displays multiple views of $Y$, each of them being a different subset of components of $X$. 

We define a view as a subset of components of $B$. We call a complete view, a view that displays full information about $Y$. According to our definition, unless it is complete, a view does not need to display full information about $Y$.

As a motivation in favor of having multiple complete views, we propose an analogy with bagging \cite{Breiman96b}. The data set $\mathcal{D}$ can be split in multiple ones according to each view. They are used to learn a set of predictors that are aggregated to form a more stable one.

\subsection{A measure of difference between views}

We suppose that, to be useful, each views have to display information about $Y$ in different ways. We propose to measure such difference with interaction information \cite{McGill54}. Let consider two views $\mathcal{I}^+$ and $\mathcal{I}^-$ being subsets of components of $B$. 
 Suppose that these two views are duplicates, such that $\mathcal{I}^+=\mathcal{I}^-$. Then for a feature $B_i \in \mathcal{I}^+$ from one view, there exists another feature $B_j \in \mathcal{I}^-$ from other view, such that $B_i=B_j$. These two features expresses the same information about $Y$. This can be characterized by $\mathbf{I}(B_i,Y|B_j)=\mathbf{I}(B_j,Y|B_i)=0$. Observation of $B_j$ (respectively $B_i$), vanishes the mutual information of $B_i$ (respectively $B_j$) with $Y$. Now suppose  $\mathcal{I}^+\cap\mathcal{I}^-=\emptyset$ such that for any features $B_i \in \mathcal{I}^+$ and $B_j \in \mathcal{I}^-$, $B_i$ and $B_j$ do not display the same information about $Y$. This can be characterized by the relations $\mathbf{I}(B_i,Y|B_j)=\mathbf{I}(B_i,Y)$ and $\mathbf{I}(B_j,Y|B_i)=\mathbf{I}(B_j,Y)$. Observation of one feature does not alter the mutual information of the other one. More generally, $B_i$ does not display the same information about $Y$ than any subset $\mathcal{I}\subset \mathcal{I}^-$, if $\mathbf{I}(B_i,Y|\mathcal{I})=\mathbf{I}(B_i,Y)$. These considerations lead us to define a measure of difference between views by measuring average interaction that have any component with up to $n$ other components and $Y$ :
$$C_B(n):=\frac{1}{m}\sum_{i}\sum_{j=1}^n\frac{1}{|\Pi^i_j|}\sum_{\mathcal{I}\in \Pi^i_j} \mathbf{I}(B_i,\mathcal{I},Y)$$
where $\mathbf{I}(B_i,\mathcal{I},Y) = \mathbf{I}(B_i,Y|\mathcal{I})-\mathbf{I}(B_i,Y)$ is the interaction information between $B_i$, $Y$ and $\mathcal{I}$. Negativity of $C_B(n)$ indicates the presence of redundancy between components. Positivity indicates presence of synergy because information about $Y$ cannot be obtained by disjoint observations of components. We shall see that there exists a relation between conditional independence of components and interaction information. Since any permutation of variables does not change the interaction information : $$\mathbf{I}(B_i,\mathcal{I},Y)=\mathbf{I}(\mathcal{I},B_i,Y)=\mathbf{I}(\mathcal{I},Y,B_i)=...$$ We can write :
$$\mathbf{I}(B_i,\mathcal{I},Y)=\mathbf{I}(B_i,\mathcal{I}|Y)-\mathbf{I}(B_i,\mathcal{I})$$
Under hypothesis of independence of components conditionally to $Y$ we have $\mathbf{I}(B_i,\mathcal{I}|Y)=0$, and consequently interaction cannot be positive.

\section{Information profile}
\label{profile}
We propose to define an information profile for the random vector $B$ that summarizes the mutual information of subsets of its components with $Y$ and that allows to compute their average interaction information. We define the information profile by the following function :
$$f_B(n) :=  \frac{1}{m} \sum_i \frac{1}{|\Pi^i_n|}\sum_{\mathcal{I} \in \Pi^i_n} \mathbf{I}(B_i,Y|\mathcal{I})$$ 
$f_B(n)$ represents the average mutual information between one component and $Y$ when $n$ other components are observed.


Information profile allows to get the average mutual information between $n$ components and $Y$ : $$F_B(n) := \frac{1}{|\Pi_n|}\sum_{\mathcal{I} \in \Pi_n} \mathbf{I}(\mathcal{I},Y)= \sum_{j=0}^{n-1} f_B(j)$$
And consequently, the quantity $\frac{m}{n}$ is an estimation of the number of non-intersecting views displaying $F(n)$ bit of information about $Y$. 


\subsection{Information profile and interaction information}

The average information interaction between any component, any other $n$ components, and $Y$, can be written :
$$\frac{1}{m} \sum_i \frac{1}{|\Pi^i_n|} \sum_{\mathcal{I} \in \Pi^i_n} \mathbf{I}(B_i,\mathcal{I},Y)$$
which can be simply computed with a difference using information profile  : $f_B(n)-f_B(0)$

This allows to compute $C_B(n)$ :
$$C_B(n) = \sum_{j=1}^{n} (f_B(j)-f_B(0))$$


\subsection{Estimation of information profile}
\label{estimate_profile_section}

Computing $f_B$ is not tractable because of sum over elements of $\Pi^i_n$ and computation of mutual information $\mathbf{I}(B_i,Y|\mathcal{I})$. We propose the following estimation :
$$\hat{f}_B(n)=  \frac{1}{m} \sum_i \frac{1}{|\mathcal{S}^i_n|}\sum_{\mathcal{I} \in \mathcal{S}^i_n} \hat{\mathbf{I}}(B_i,Y|\mathcal{I})$$
where $\mathcal{S}^i_n$ is a set of samples from the set $\Pi^i_n$. $\hat{\mathbf{I}}(B_i,Y|\mathcal{I})$ is an estimation of the mutual information using the following estimation of conditional entropy :

$$\hat{\mathbf{H}}(B_i|\mathcal{I},Y) = \frac{1}{|\Gamma|}\sum_{(b,y) \in \Gamma} \mathbf{H}(B_i|b,y)$$

where $\Gamma$ is a set of samples from $P(\mathcal{I},Y|\theta)$. We can sample from $P(\mathcal{I},Y|\theta)$ by randomly picking an example $(x,y)\in\mathcal{D}$, then sample from $P(\mathcal{I}|x,\theta)$, which is easy if we have \eqref{inde}.

Similarly, we can estimate $\mathbf{H}(B_i|\mathcal{I})$, and then get an estimation of mutual information : $$\hat{\mathbf{I}}(B_i,Y|\mathcal{I}) = \hat{\mathbf{H}}(B_i|\mathcal{I}) - \hat{\mathbf{H}}(B_i|\mathcal{I},Y)$$

To evaluate the quality of estimation $\hat{f}_B$, we compared it with the computable information profile of a binary random vector $B^\alpha$, which distribution is defined and parametrized by $\alpha$ as following.
We assume the conditional independence of components with respect to $Y$ : 
$$P(B^\alpha|Y) = \prod_{j} P(B^\alpha_j|Y)$$
We also suppose that $Y$ is binary and uniform, such that $P(Y=1)=P(Y=0)=\frac{1}{2}$. The following hypotheses complete to define distribution of $B^\alpha$ :
\begin{eqnarray*}
\forall j, P(B^\alpha_j=0|Y=0)=\alpha \\
\forall j, P(B^\alpha_j=1|Y=1)=\alpha 
\end{eqnarray*}
with $\frac{1}{2}\leq \alpha \leq 1$. 

The information profile of $B^\alpha$ is computable because we have :
\begin{equation}
\label{B_prop}
|\mathcal{I}|=|\mathcal{J}| \Rightarrow \mathbf{I}(B^\alpha_i,Y|\mathcal{I})=\mathbf{I}(B^\alpha_i,Y|\mathcal{J})
\end{equation}



\begin{figure*}[!ht]
\vskip 0.2in
\begin{center}
\subfigure[ Comparison between $f_{B^\alpha}$ and its estimate $\hat{f}_{B^\alpha}$, for $\alpha\in \{0.57,0.58,0.6\}$.]{
\includegraphics[width=200pt]{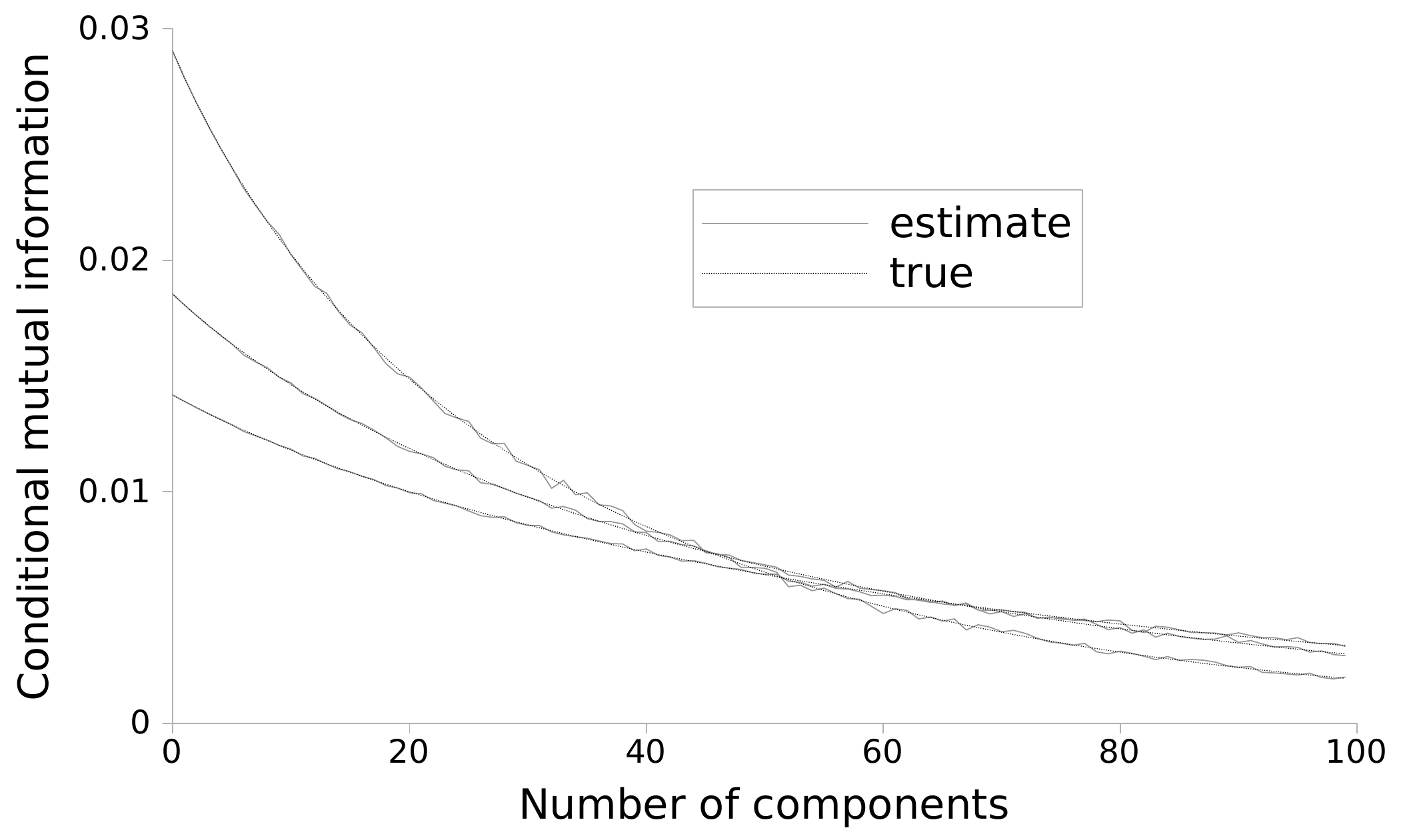}
\label{test_estimation}
}\quad
\subfigure[Information profiles estimates $\hat{f}_B$ computed on the train set.]{
\includegraphics[width=200pt]{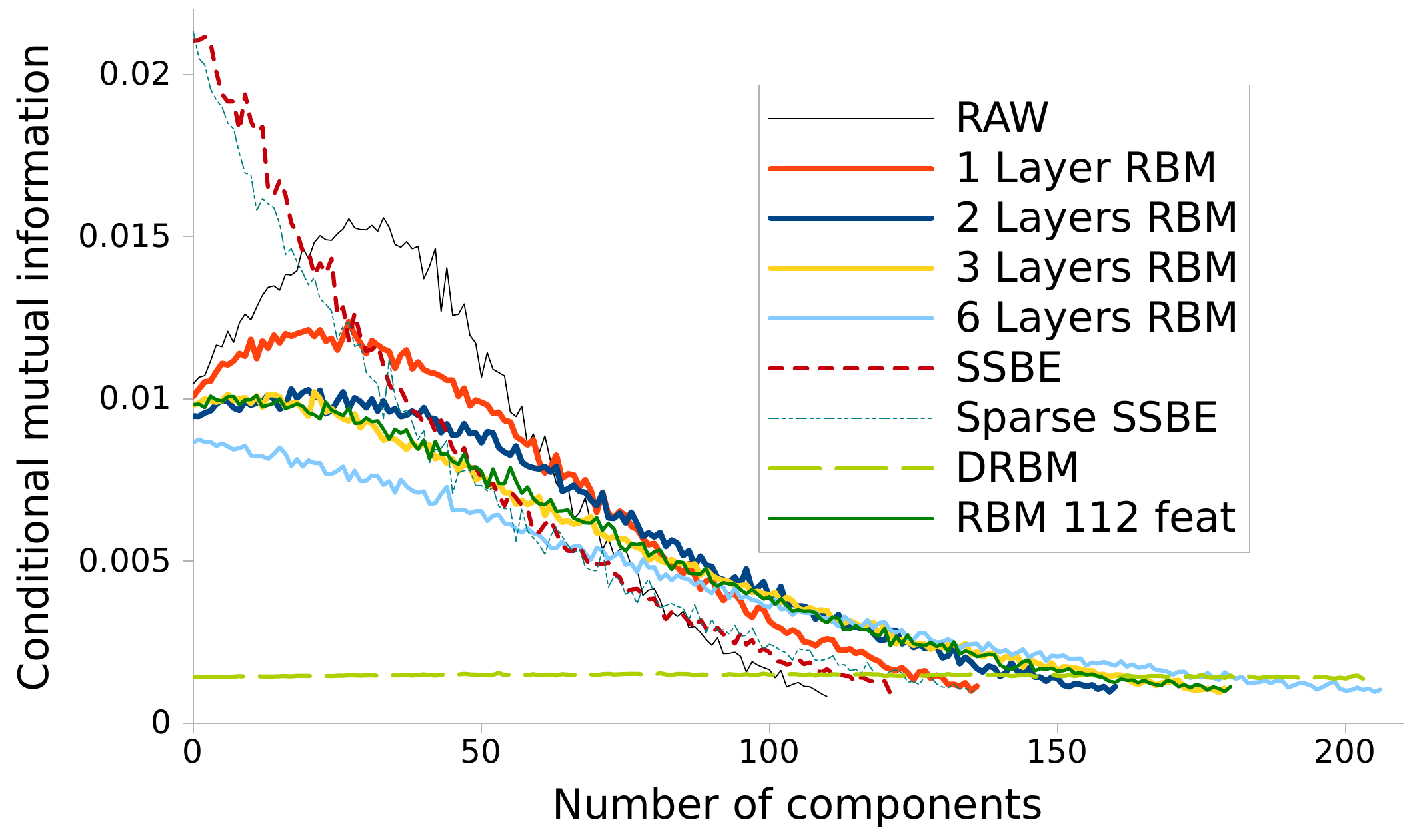}
\label{profiles}
}
\subfigure[For each size $n$ of subset of components from the representation, the average mutual information $\hat{F}_B(n)$ (on the x-axis) versus the number of non-intersecting views $\frac{m}{n}$ (on the y-axis).]{
\includegraphics[width=200pt]{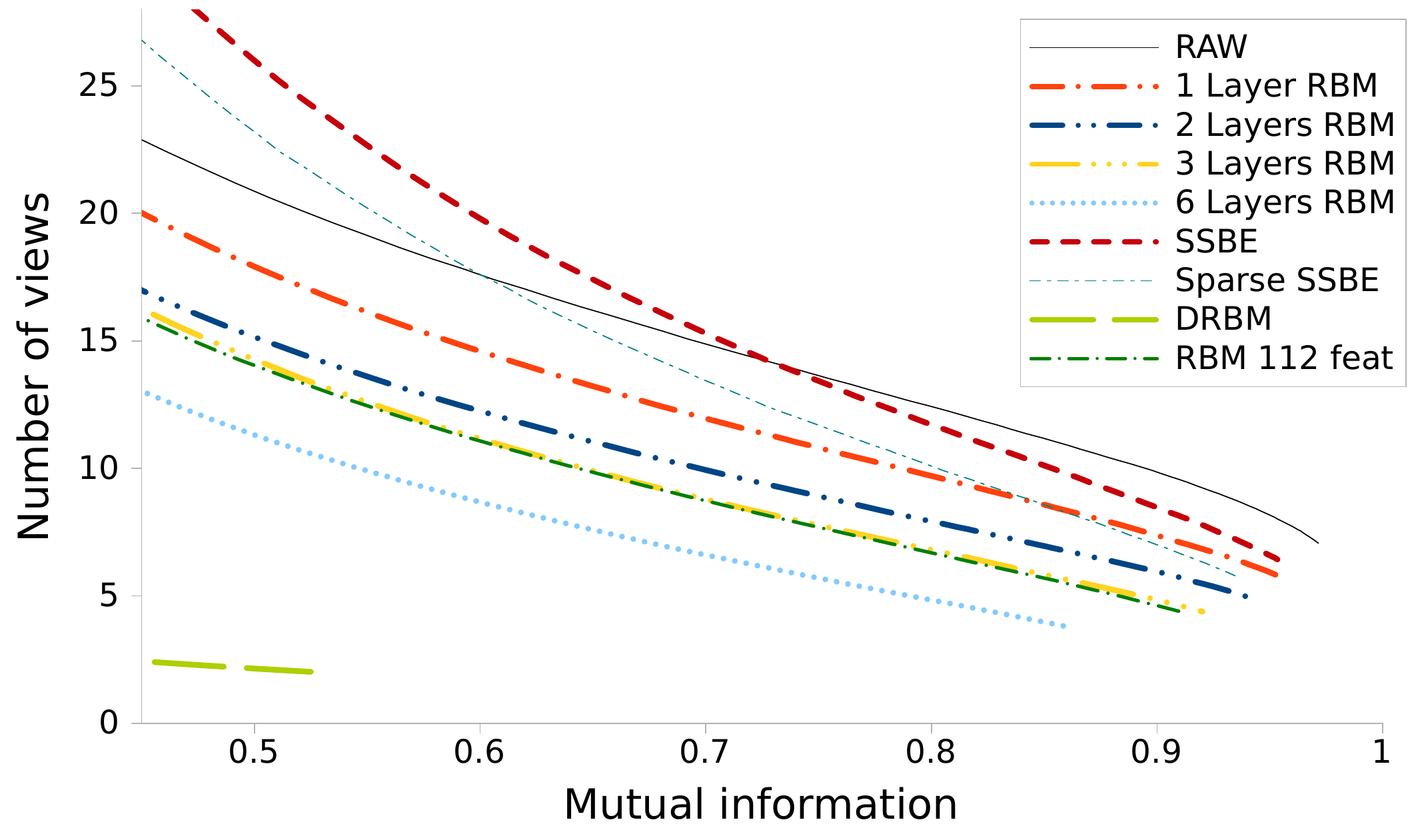}
\label{views}
}\quad
\subfigure[For each size $n$ of subset of components from the representation, the average mutual information $\hat{F}_B(n)$ (on the x-axis) versus the average interactions $\hat{C}_B(n)$ (on the y-axis).]{
\includegraphics[width=200pt]{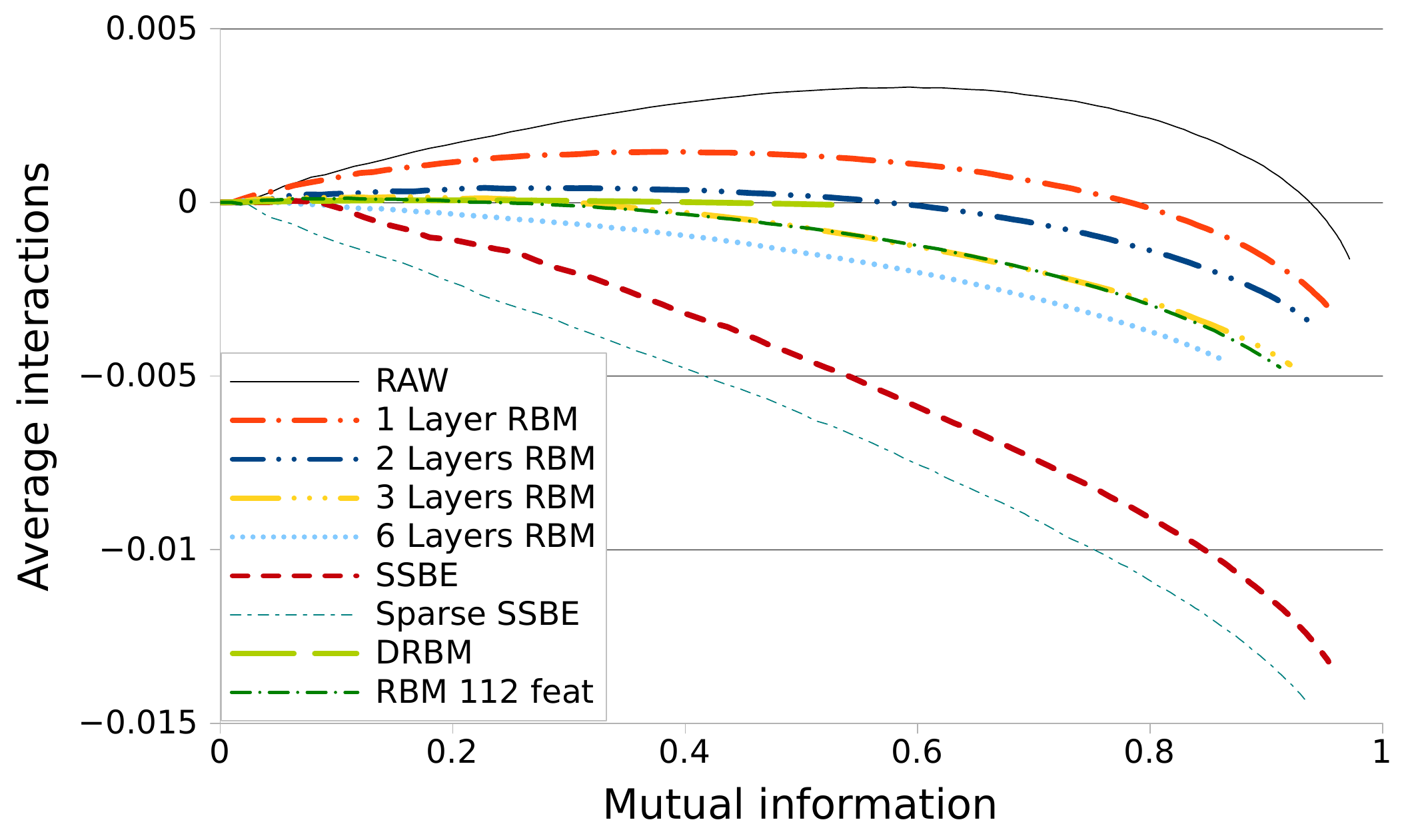}
\label{interactions}
}
\subfigure[For each size $n$ of subset of components from the representation, the average mutual information $\hat{F}_B(n)$ (on the x-axis) versus the average normalized conditional total correlation $\hat{D}_B(n)$ (on the y-axis).]{
\includegraphics[width=200pt]{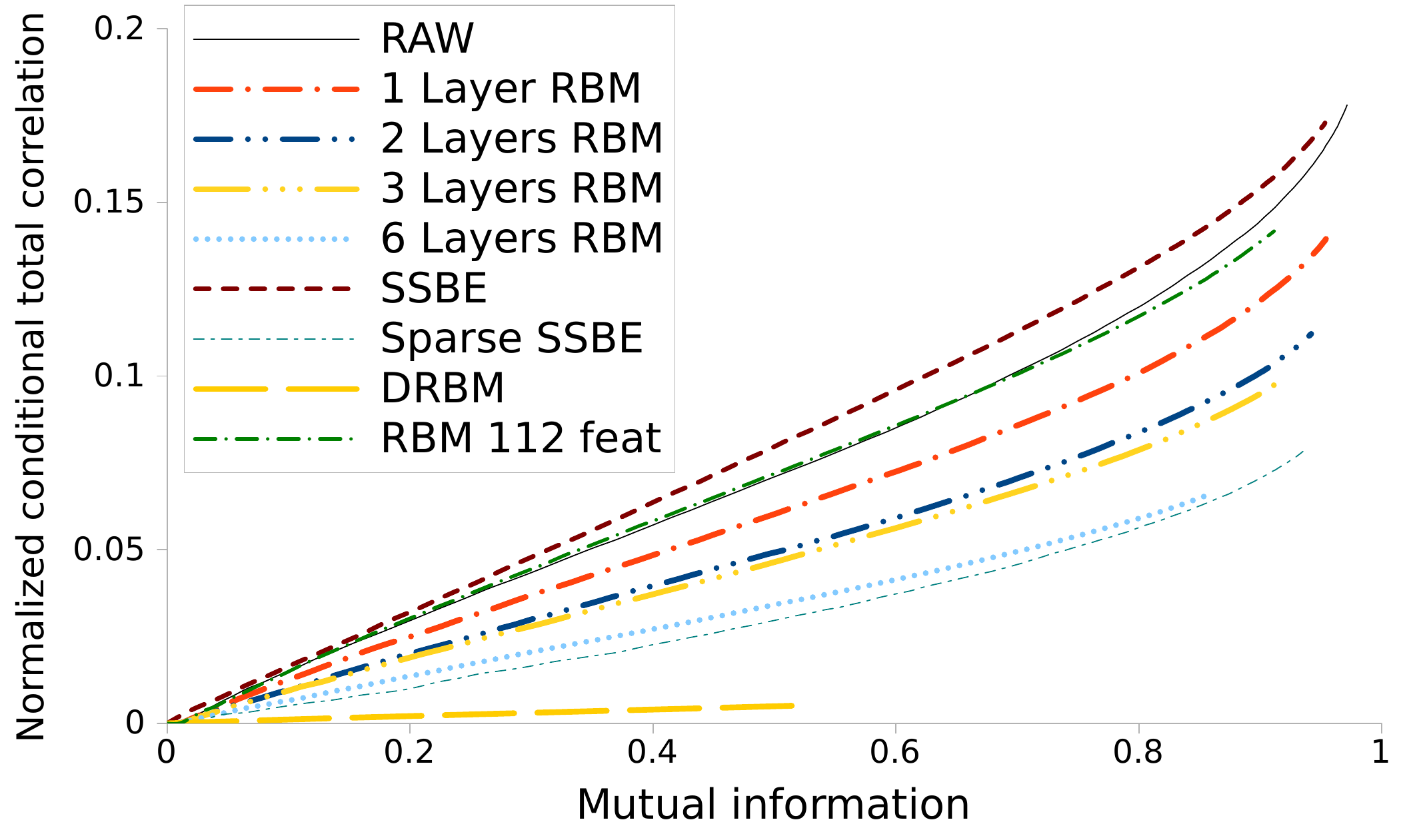}
\label{correlations}
}
\caption{Unsupervised models manage to conserve views from the input distribution, whereas supervised models do not.}
\end{center}
\vskip -0.2in
\end{figure*} 

\begin{figure*}[!ht]
\vskip 0.2in
\begin{center}
\subfigure[Conjoint observation with the figure \ref{size_interactions} show that larger representations allow to conserve more views from input distribution.]{
\includegraphics[width=200pt]{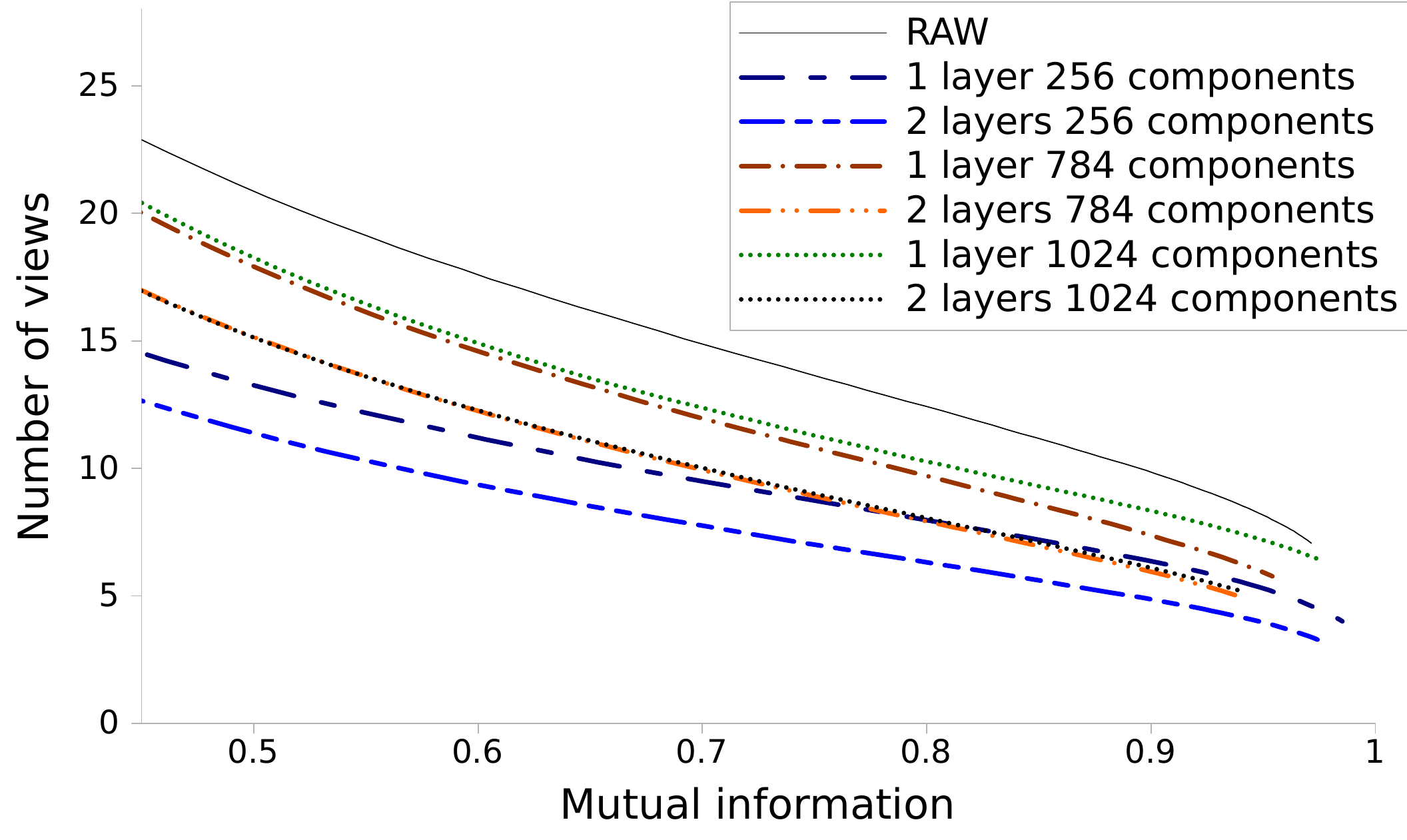}
\label{size_views}
}\quad
\subfigure[Larger representations manage to learn non interacting features. Smaller representations, either do not manage to solve positive interactions or learn redundant features.]{
\includegraphics[width=200pt]{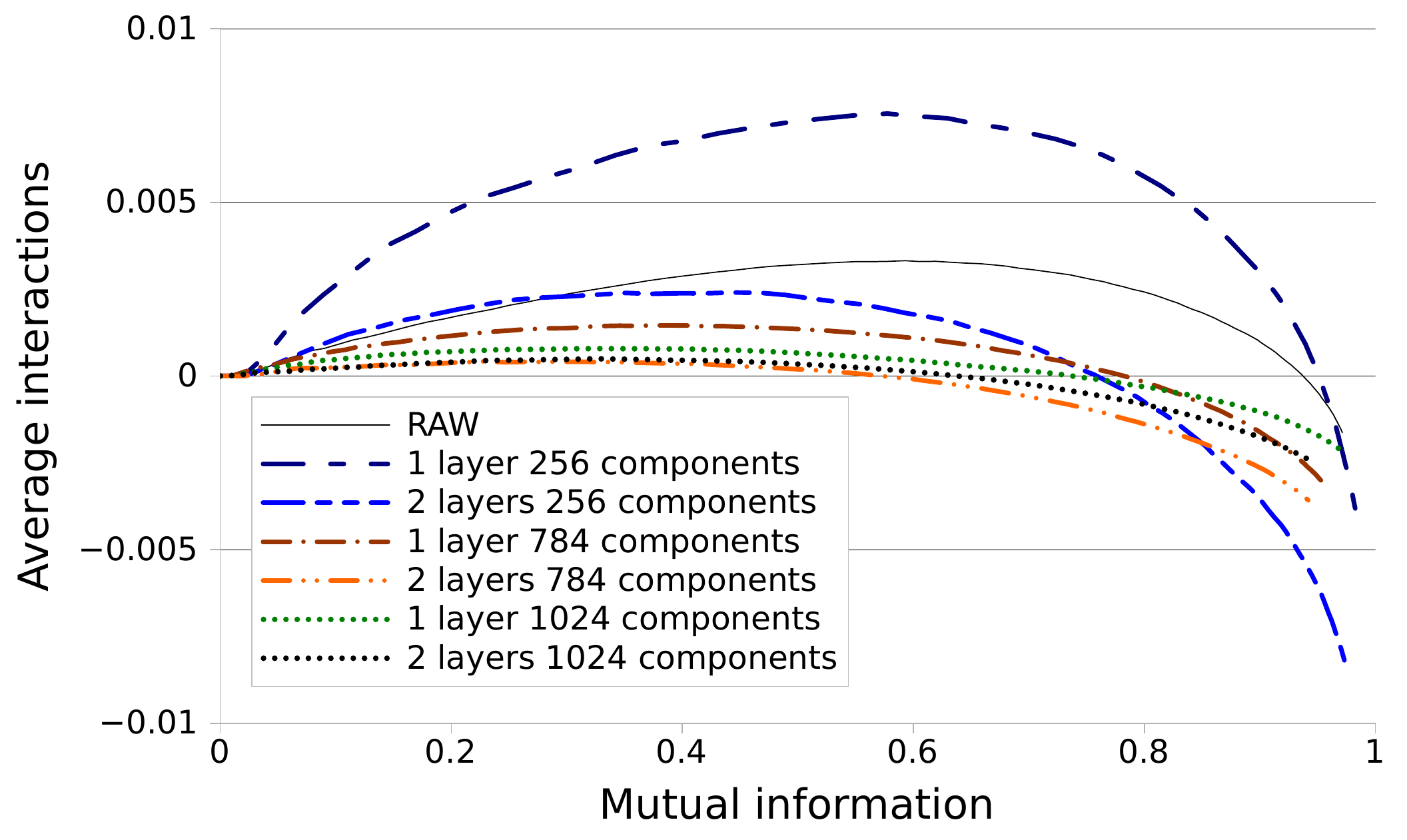}
\label{size_interactions}
}\quad
\subfigure[Larger representations manage to have low conditional total correlation, while smaller representations do not.]{
\includegraphics[width=200pt]{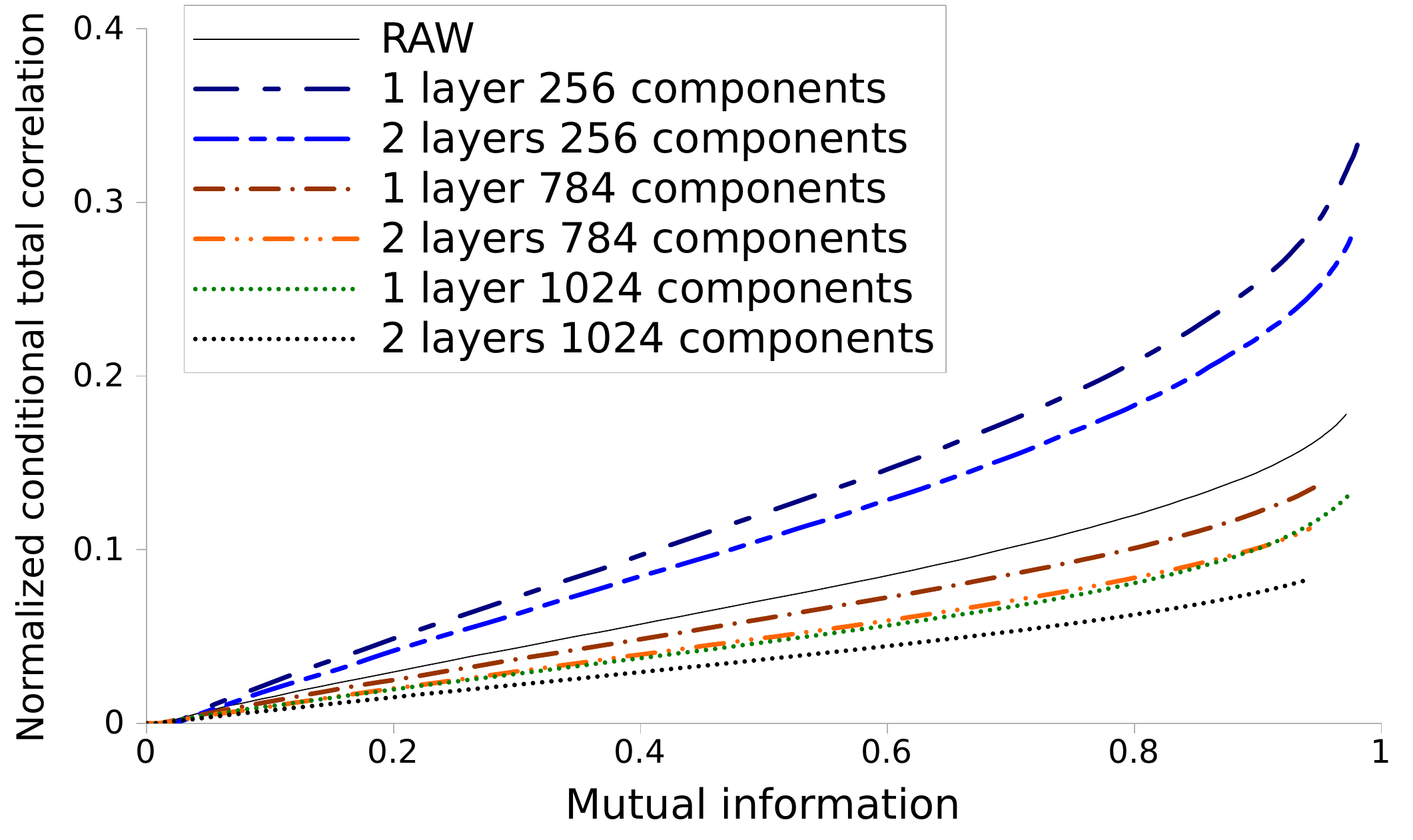}
\label{size_cor}
}
\caption{Effects of the size of the representation on the quantity of preserved views and disentanglement of supervision.}
\end{center}
\vskip -0.2in
\end{figure*}

Figure \ref{test_estimation} shows the accuracy of estimate $\hat{f}_{B^\alpha}(n)$ for various values of $\alpha$. The number of samples is $|\Gamma|=32$ and $|\mathcal{S}^i_n|=100$. The estimate is accurate, however more variance is expected for distributions that do not satisfy \eqref{B_prop}.

We note $\hat{F}_B$ the estimate of $F_B$ obtained using $\hat{f}_B$.

\subsection{Estimation of $D_B(n)$ }

We use estimates of conditional entropies from section \ref{estimate_profile_section} to compute an estimate $\hat{D}_B(n)$ of $D_B(n)$ :
$$\hat{D}_B(n)=  \frac{1}{m} \sum_{j=1}^n \sum_i \frac{1}{|\mathcal{S}^i_j|}\sum_{\mathcal{I} \in \mathcal{S}^i_j}\left( \hat{\mathbf{H}}(B_i|Y)-\hat{\mathbf{H}}(B_i|\mathcal{I},Y)\right)$$

\section{Experiments}
\label{experiments}
We performed experiments on a variation of the well know MNIST data set, in which the digit labels are grouped in two classes depending on their parity. This data set represents a variable $Y$ that has one bit of entropy. We learned representations with both supervised and unsupervised models. For the former case, we stacked RBMs, for the latter we used a discriminative RBM (DRBM) \cite{DRBM} and a stochastic supervised binary encoder (SSBE). 

%

\paragraph{Stochastic supervised binary encoder :} A SSBE is a supervised model designed to learn multiple views. This model uses a learning objective function that aims to maximize the mutual information of $Y$ with fixed-size subsets of components of the representation. We use the following model : $$P(B|X,\theta=\{W,c\})=\sigma(W.X+c)$$ where $\sigma$ is the sigmoid function\footnote{$\sigma(x)=\frac{1}{1+e^{-x}}$} , $W$ is a\footnote{we note $m_X$ the dimensionality of $X$} $m\times m_X$ matrix of weights, and $c$ a $m$-dimensional vector of biases. The learning objective aims to maximize mutual information $\mathbf{I}(\mathcal{I},Y)$ for any subset of components $\mathcal{I}\in \Pi_n$, its objective function is written :
\begin{eqnarray*}\label{obj} \theta^* & = &\text{arg}\max_{\theta} \sum_{\mathcal{I} \in \Pi_n} \mathbf{I}(\mathcal{I},Y|\theta)\\
& = & \text{arg}\min_{\theta} \sum_{\mathcal{I} \in \Pi_n} \mathbf{H}(Y|\mathcal{I},\theta)
\end{eqnarray*}
The gradient over parameters $\theta$ is intractable, but we can compute an estimation with samples from the distribution $P(Y,B,X|\theta)=P(Y,X)P(B|X,\theta)$ using data set $\mathcal{D}$ and current parameters $\theta$.


Unless specifically mentioned, all representations are learned with the same size $m=784$ corresponding to the input dimension. On figures, \textit{RAW} designates the input distribution by interpreting pixels as probabilities. \textit{RBM 112 Feat} is a learned RBM from which we picked up 112 of its hidden variables, they were duplicated seven times to form a new representation. \textit{Sparse SSBE} is a SSBE learned with a sparse regularization applied on components of $B$. This is done by adding the following term to the objective function :
\[\lambda\sum_{i}\limits D_{KL}(\mathcal{B}(p)\|P(B_i|\theta))
\]
where $D_{KL}(.||.)$ designates the Kullback-Leibler divergence, $\mathcal{B}(p)$ is the Bernoulli distribution of parameter $p$. Hyper-parameters are $\lambda$ that controls the magnitude of the regularization, and $p$ that controls the sparsity level.

The figure \ref{profiles} shows information profile estimates computed on the training set\footnote{We got same profiles on the test set.}. We computed $\hat{f}_B(n)$ with increasing values of $n$ until $\hat{f}_B(n)<0.001$, this explains missing data for some models. The number of samples was $|\Gamma|=32$ and $|\mathcal{S}^i_n|=100$. We see that profile of \textit{RAW} has its maximum at around $40$ components, with a high difference $\hat{f}_B(40)-\hat{f}_B(0)$, this indicates strong positive interactions in the input distribution. These are resolved as we increase layer depth for stacked RBMs (profiles flatten).

On figures \ref{views},\ref{interactions} and \ref{correlations}, are displayed for each size $n$ of subset of components, the average mutual information $\hat{F}_B(n)$ (on the x-axis) versus (on the y-axis) : 
\begin{itemize}
\item number of non-intersecting views $\frac{m}{n}$ (figure \ref{views}),
\item the average interactions $\hat{C}_B(n)$ (figure \ref{interactions}),
\item the average normalized conditional total correlation $\hat{D}_B(n)$ (figure \ref{correlations}).
\end{itemize}

For stacked RBMs, we observe that, as the number of layer increases, the total correlation decreases (figure \ref{correlations}), indicating improvement of disentanglement of supervision. However, the figure \ref{views} shows a counterpart which is the decreasing number of views as we increase depth. This can be an explanation of why increasing further the number of layer decreases the classification performances as shown in \cite{Erhan-aistats-2010}.

The partial data for the DRBM shows that the model has likely learned only one view (figure \ref{views}). The model has retained only enough information to achieve its learning objective.

The SSBE learned without sparsity displays high conditional total correlation indicating no disentanglement of classes (figure \ref{correlations}). This is not surprising since its objective is to maximize mutual information without constraint on how this information have to be arranged. However, it appears that adding a sparsity constraint greatly reduces the total correlation. The SSBE was specially designed to learn multiple views in a supervised fashion. It has apparently succeeded as seen on figure \ref{views}, but the figure \ref{interactions} shows that they are highly redundant.

The RBM with duplicated features (\textit{RBM 112 Feat} on figures) shows, as expected, a reduced number of views compared to the original RBM (figure \ref{views}).  It has a similar profile than 3 stacked RBMs, except that it has a higher conditional total correlation.

Finally, figures \ref{size_views}, \ref{size_interactions} and \ref{size_cor} show effect of variation of the size of representation learned with stacked RBMs. We see that larger representation helps to have both non interacting components and low conditional total correlation, while smaller representation fails to solve positives interactions from the input distribution and has high conditional total correlations.

\section{Discussions and future work}
We proposed to analyze the representations learned with unsupervised pre-training under the perspective of information theory. We proposed two measures :
\begin{itemize}
\item a measure of conditional independence of components. We showed that conditional independence of components in the context of binary features enables classes to be linearly separable.
\item a measure of quantity of views of the supervision within the representation. We supposed that the input distribution displays various views about the supervision, and that learning a representation that preserves them helps to better generalize because they provide information that help to determine parameters of the model.
\end{itemize}
Experiments showed that, according to our measures, representations learned with unsupervised models succeeded to conserve views from input distribution, whereas supervised attempts failed. 


Our approach suggests a new learning objective for learning representations : disentangling supervision while trying to conserve views about supervision. Some information displayed by input may be irrelevant for the supervision,  e.g. textures or backgrounds on image recognition tasks, transferring them in the representation would be wasteful. As future work, this suggests that biasing unsupervised models by integrating the supervision signal during pre-training may help to conserve views by keeping only the relevant information and improve classification performance.




\bibliography{example_paper}
\bibliographystyle{icml2012}

\end{document}